\documentclass[10pt,twocolumn,letterpaper]{article}

\usepackage[pagenumbers]{cvpr} 

\usepackage{graphicx}
\usepackage{amsmath}
\usepackage{amssymb}
\usepackage{booktabs}
\usepackage{multirow} 

\usepackage[breaklinks,colorlinks]{hyperref}

\usepackage[capitalize]{cleveref}
\crefname{section}{Sec.}{Secs.}
\Crefname{section}{Section}{Sections}
\Crefname{table}{Table}{Tables}
\crefname{table}{Tab.}{Tabs.}

\begin{document}

\title{
Multi-View Neural 3D Reconstruction of Micro-/Nanostructures \\with Atomic Force Microscopy
}

\author{
Shuo Chen$^1$ \hspace{1cm}
Mao Peng$^2$ \hspace{1cm}
Yijin Li$^1$ \hspace{1cm}
Bing-Feng Ju$^2$ \hspace{1cm}
Hujun Bao$^1$\\
Yuan-Liu Chen$^{2*}$ \hspace{1cm}
Guofeng Zhang$^{1*}$ \vspace{0.5em}\\
\textnormal{\fontsize{11pt}{14pt}\selectfont $^1$State Key Lab of CAD\&CG, Zhejiang University, Hangzhou 310058, China} \\
\textnormal{\fontsize{11pt}{14pt}\selectfont $^2$State Key Lab of Fluid Power\&Mechatronic Systems, Zhejiang University, Hangzhou 310027, China} \\
\textnormal{\tt\small \{chenshuo.eric, pengmao99, eugenelee, mbfju, baohujun, yuanliuchen, zhangguofeng\}@zju.edu.cn}
}
\maketitle

\def\thefootnote{$\ast$}\footnotetext{corresponding authors}

\begin{abstract}

Atomic Force Microscopy (AFM) is a widely employed tool for micro-/nanoscale topographic imaging. However, conventional AFM scanning struggles to reconstruct complex 3D micro-/nanostructures precisely due to limitations such as incomplete sample topography capturing and tip-sample convolution artifacts. Here, we propose a multi-view neural-network-based framework with AFM (MVN-AFM), which accurately reconstructs surface models of intricate micro-/nanostructures. Unlike previous works, MVN-AFM does not depend on any specially shaped probes or costly modifications to the AFM system. To achieve this, MVN-AFM uniquely employs an iterative method to align multi-view data and eliminate AFM artifacts simultaneously. Furthermore, we pioneer the application of neural implicit surface reconstruction in nanotechnology and achieve markedly improved results. Extensive experiments show that MVN-AFM effectively eliminates artifacts present in raw AFM images and reconstructs various micro-/nanostructures including complex geometrical microstructures printed via Two-photon Lithography and nanoparticles such as PMMA nanospheres and ZIF-67 nanocrystals. This work presents a cost-effective tool for micro-/nanoscale 3D analysis.

\end{abstract}

\section{Introduction}
\label{sec:intro}
The investigation of the three-dimensional (3D) structure plays a vital role in nanotechnology research,  encompassing areas like nanofabrications~\cite{nanofabrication, nanofabrication2}, nanorobots~\cite{nanorobot, nanorobot2}, and nanomedicines~\cite{particle_cell, Nanomedicines}, given its critical relevance to the functional properties of micro-/nanoscale objects.
Currently, the Scanning Electron Microscope (SEM)~\cite{SEM} is a prevalent tool for observing the 3D geometry of micro-/nanostructures.
This technique involves irradiating the sample with an electron beam and capturing a 2D image by detecting the intensity of secondary electrons emitted from the sample surface.  
Despite its widespread use and multiple advantages, SEM is a destructive method~\cite{SEM_damage}, requires a vacuum environment, and cannot provide accurate height information in the images.
In contrast, the Atomic Force Microscope (AFM)~\cite{AFM_root} acquires precise height information of the sample surface through the forces between its probe and the sample. Moreover, AFM can operate in various environments, is insensitive to the sample material, and is non-destructive.

Nonetheless, conventional AFM comes with its own set of challenges. 
One primary limitation is that conventional AFM can only capture 2.5D information instead of a complete 3D representation of the sample because the position feedback in conventional AFM systems is confined to the vertical direction~\cite{AFM_root}.
Another significant challenge is the issue of tip-sample convolution~\cite{AFM_image_artifacts}. 
This phenomenon arises from geometrical interactions between the AFM tip and the surface features of the sample~(Supplementary \cref{fig: afm_limit}).
These interactions often lead to artifacts~\cite{AFM_image_artifacts, AFM_image_artifacts_bacterial, AFM_image_artifacts_thin_film} in the scanning results that are inherently difficult to differentiate from the actual sample geometry.
Such limitations impede the effective use of AFM to investigate intricate 3D micro-/nanostructures and catalyze the development of advanced AFM technologies, i.e., 3D-AFM.

The advancement of 3D-AFM technology predominantly follows two distinct trajectories. 
The first approach involves the design of specialized probe shapes aimed at enabling the measurement of structures that are inaccessible with conventional AFM scanning. 
As an example, critical dimension AFM (CD-AFM)\cite{CD-AFM, CD-AFM_2006}, currently a prevalent method for semiconductor structures, utilizes flared tips that enable lateral dithering in addition to the vertical oscillation of the cantilever. 
These designs equip the AFM with the capability to image not only vertical but also undercut sidewall features of samples.
Additionally, there are other creative designs of probes, such as the introduction of hinge structures~\cite{CD-AFM_probe}, orthogonal cantilevers~\cite{OCP}, and probes made of carbon nanotube with high aspect ratios\cite{CNT_probes}. 
However, the extra cost and complexity of manufacturing special probes and customized AFM scanning systems present substantial challenges to the widespread adoption of these methods.

Another common technique for 3D-AFM is the practice of tilting either the probe~\cite{Park_System_corporation, tilt_icp1, Scanning_Density, Dual_Probes_AFM1, Dual_Probes_AFM2} or the sample~\cite{tilt_icp3, tilt_icp2} to scan micro-/nanostructures from multiple directions. 
These methods avoid the necessity for specialized flared tips, instead relying on integrating multiple scans into a complete 3D model.
As a result, the effectiveness relies on the precision of the data stitching of multiple scans.
Historically, previous methods~\cite{Park_System_corporation, tilt_icp1, Scanning_Density, Dual_Probes_AFM1, Dual_Probes_AFM2, tilt_icp3, tilt_icp2} predominantly apply to simple, well-defined structures, such as gratings. 
The grating's relatively straightforward structure facilitates the manual removal of artifacts resulting from tip-sample convolution and simplifies the problem of the tilting method.
However, significant challenges arise when the tilting method is applied to micro-/nanostructures of unknown and intricate shapes, such as those created by Two-photon Lithography (TPL)~\cite{TPL_origin_nature, TPL_2023, 3d-print_functional_microrobots,3d-print_microrobots} or comprised of diverse nanoparticles~\cite{particle_cell, Nanomedicines, metal_nanocrystals2}—a scenario frequently encountered in nanotechnology research.
Firstly, the complex surface geometries of these structures make it difficult to manually identify and remove artifacts from the AFM images.
Secondly, the existence of these unremoved artifacts in the scans impedes the accuracy of the stitching process. 
Thirdly, owing to the complex overlapping relationships among multi-view data, simply stitching of these data is insufficient for the construction of a clear and accurate 3D model.

In this study, we propose MVN-AFM, a framework that is able to reconstruct the 3D surface model of a wide range of complex-shaped micro-/nanostructures without any specially shaped probes or costly modifications to the AFM system.
Our framework leverages the concept of tilting samples, but we extend its application for complex structures beyond the limitations of existing methods.
Specifically, we first propose an iterative optimization algorithm to automatically remove AFM artifacts and improve the alignment accuracy of multi-view data from intricate micro-/nanostructures.
Subsequently, in order to reconstruct the 3D model of these structures by multi-view AFM data, we draw inspiration from multi-view depth fusion techniques~\cite{Kinectfusion, classical_recon1, classical_recon2, Neural_survey, neuralfusion,bvn-fusion} in computer vision. 
We introduce the neural implicit surface reconstruction methods~\cite{nerf, unisurf, IDR, volsdf, imap, Neus}, the recent advance in this field, to utilize a neural network to represent the 3D model of micro-/nanostructures. By employing differentiable volume rendering to train the neural implicit function with multi-view AFM data supervision, we fuse the multi-view scanning results into an accurate and comprehensive 3D model. 
Furthermore, we conduct extensive experiments to evaluate the capabilities of the MVN-AFM framework. In detail, we utilize the TPL technique to fabricate various 3D microstructures with distinct geometrical characteristics and prepare specimens of commonly used nanoparticles, including polymethyl methacrylate (PMMA) nanospheres~\cite{PMMA_drug, PMMA1, PMMA2} and Zeolitic imidazolate
framework (ZIF)-67 nanocrystals~\cite{ZIF-67_application, synthesis_ZIF67, ZIF67_Electrochemically, ZIF67_Catalyst}. MVN-AFM effectively eliminates artifacts present in raw AFM images and successfully reconstructs not only the overall shape but also specific hidden details that are not discernible in conventional AFM scans. The ability of MVN-AFM to provide detailed and accurate 3D reconstructions of a broad spectrum of micro-/nanostructures, coupled with its low implementation cost, positions it as a potentially valuable tool in nanotechnology research.

\section{Results}
\label{sec:results}

\subsection{Pipeline of MVN-AFM}

\begin{figure*}[h] 
    \centering
    \includegraphics[width=\linewidth]{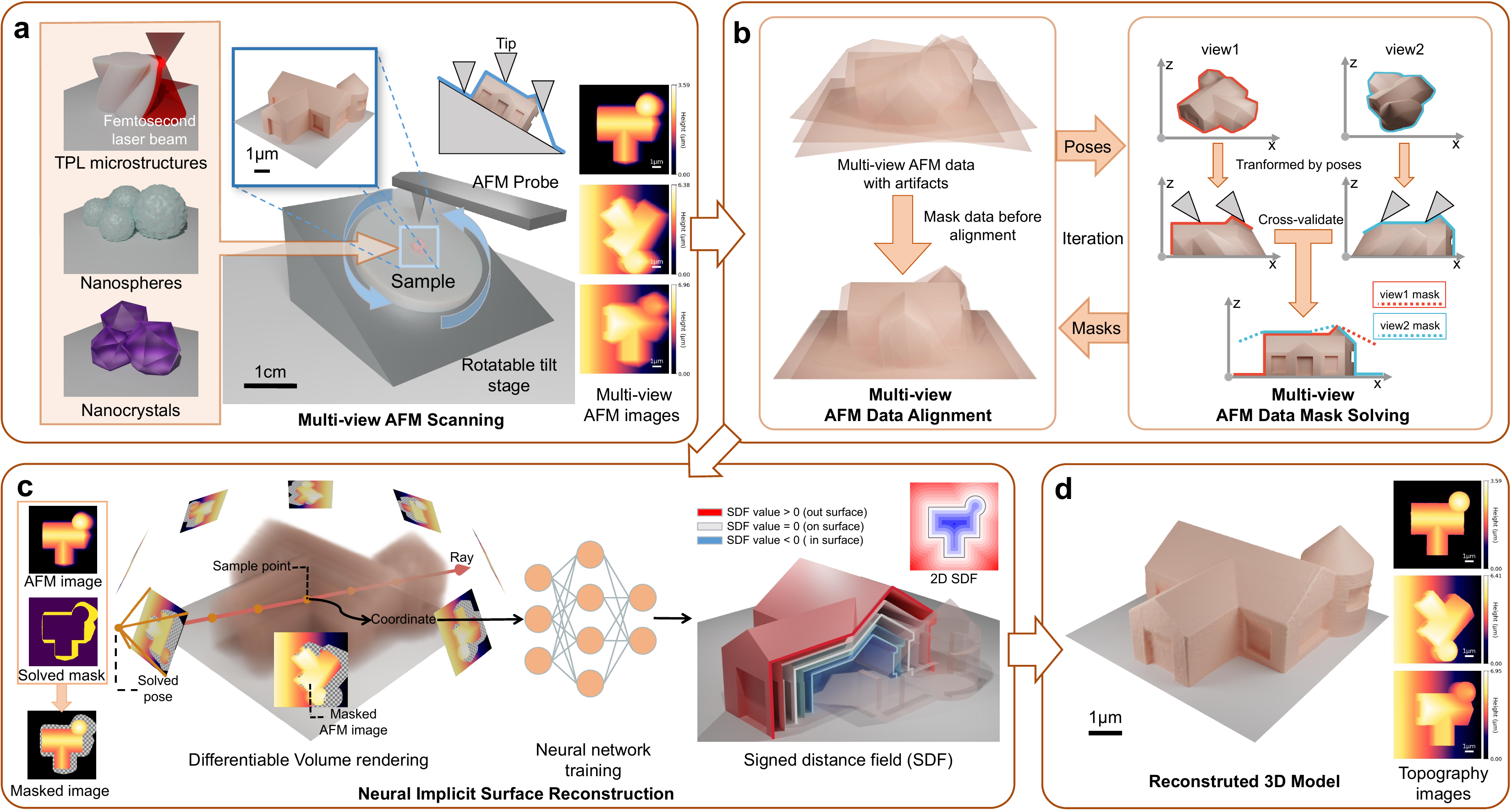}
    \caption{ \textbf{The pipeline of MVN-AFM.} \textbf{a} First, we place various micro-/nanostructures on a rotatable tilt stage. Second, we rotate the turntable and measure the vertical heights by a conventional AFM, resulting in a set of multi-view AFM images with many artifacts. \textbf{b} Input raw AFM images with artifacts and iterate two sub-steps. In the data alignment process, data judged as artifacts are eliminated before alignment, and the poses of multi-view images are updated. In the mask-solving process, the solved poses transform the multi-view data, and the data consistency is cross-validated to solve the mask of artifacts.
    \textbf{c} The posed and masked multi-view AFM images are used to train a neural network representing a signed distance field in space by the differentiable volume rendering technique (Supplementary~\cref{fig: network_pipeline}). \textbf{d} The 3D surface model extracted from the signed distance field, and corresponding topography images without artifacts.}
    \label{fig: pipeline}
    \vspace{-1em}
\end{figure*}

The objective of MVN-AFM is to construct a generalized process that can be used for 3D reconstruction of unknown-shaped complex micro-/nanostructures based on multi-view AFM scanning data, relying only on conventional AFM systems and standard probes.
MVN-AFM consists of three main steps (\cref{fig: pipeline}): \textit{Multi-view AFM Scanning}, \textit{Data Alignment and Mask Solving}, and \textit{Neural Implicit Surface Reconstruction}.

The step of \textit{Multi-view AFM Scanning} (\cref{fig: pipeline}a)  captures multi-view AFM images of 3D micro-/nanostructures, providing essential geometric information for the subsequent reconstruction process.
Previous tilting methods~\cite{Park_System_corporation, tilt_icp1, Scanning_Density, Dual_Probes_AFM1, Dual_Probes_AFM2, tilt_icp3, tilt_icp2} acquire complete geometric information with only two AFM scans towards each sidewall of the grating. However, in nanotechnology research, the prior knowledge of the sample's shape and orientation is often unknown.
To address this, we design a standardized AFM scanning process that is independent of the sample's shape. 
This process aims to comprehensively acquire the surface geometric information of unknown and complex-shaped structures.
We use the sample-tilting approach to collect multi-view data, thus avoiding modifications to the mechanics of conventional AFM.
For this purpose, we designed a rotatable stage with a tilt angle (Supplementary \cref{fig: rotate_base}). 
We carefully designed the size of the whole stage so that it can be used in the limited activity space of a commercial AFM without any collision.
The sample is placed on a turntable in the center of the stage so that multi-view scans around the sample can be acquired as it rotates to different directions.

In the step of \textit{Data Alignment and Mask Solving}~(\cref{fig: pipeline}b), we iteratively align multi-view AFM data to a unified coordinate system and remove the artifacts in AFM images. 
A critical process in the tilting method involves establishing the spatial relationship among multi-view data. This requires determining a coordinate transformation, denoted as pose $T$, to align data from different views within the same coordinate system (Supplementary~\cref{fig: coordinate}). 
To achieve this, some methods~\cite{Park_System_corporation, Dual_Probes_AFM1, Dual_Probes_AFM2} employ designs with high-cost components to enable precise control of probe scanning direction, which allows direct access to $T$.
Others~\cite{tilt_icp1,tilt_icp2,tilt_icp3} utilize the Iterative Closest Point (ICP) algorithm~\cite{ICP} to solve $T$ by minimizing the distance between AFM data points. The ICP algorithm relies heavily on data free from artifacts that do not represent the actual sample shape.
Consequently, these previous methods~\cite{tilt_icp1,tilt_icp2,tilt_icp3} manually remove highly recognizable artifacts from AFM data of simple structures before using the ICP algorithm.
However, for multi-view images obtained by a conventional AFM system on intricate structures, there are two challenges: eliminating artifacts and solving for the pose. 
Here, we define the label of whether each data point is an artifact as a latent variable, mask $M$.
To simultaneously solve for $T$ and $M$, we propose an iterative EM-like algorithm~\cite{EM_intro, EM_root}.
Initially, we consider all AFM data artifact-free, i.e., $M_0$ is all zeros, and directly apply the ICP algorithm to obtain a set of coarse poses, $T_0$.
In the E-step, we project multi-view data onto each other using $T_{i-1}$ from the previous iteration $i-1$. We then conduct cross-validation of the projected data to identify areas of inconsistency in multi-view data. These regions are then labeled as ones, and we obtain the updated $M_i$.
The motivation for the cross-validation is that artifacts vary with the probe-sample angle, so they are inconsistent in multi-view data. In contrast, the sample's geometric surface remains consistent across different views, regardless of the probe scanning directions.
In the M-step, we erase the artifact through $M_i$ and apply the ICP algorithm again to compute the updated $T_i$.
Iterating the EM steps, the data filtered out of most artifacts by $M$ yields a more precise $T$. The improved $T$ also makes the cross-validation accurate. Two steps are iteratively performed to enhance each other.
After $k$ iterations, we obtain the accurate pose $T_k$ and the artifact mask $M_k$ for each viewpoint of the AFM data.

The step of \textit{Neural Implicit Surface Reconstruction} (\cref{fig: pipeline}c) utilizes the aligned and masked AFM data to train an implicit function represented by a neural network and extract the final 3D surface model of micro-/nanostructures from the network.
Specifically, we follow previous work~\cite{SDF_Loss} and model the geometry surface of the sample as a neural network encoded Signed Distance Field (SDF): $s(x; \theta): \mathbb{R}^3 \rightarrow  \mathbb{R}$, where $x$ denotes a 3D position and $\theta$ is the parameters of a Multilayer Perceptron (MLP).
The SDF defines a scalar field where each point in space is associated with the shortest distance to a surface. This distance is positive if the point is outside the surface and negative if it is inside.  
Previously, the neural implicit surface reconstruction methods~\cite{Neus} were developed for posed images from cameras in the macroscopic world, not for nanotechnology and AFM data applications.
To adapt this method to our reconstruction process, we convert AFM images into depth maps as captured by virtual orthogonal cameras (Supplementary~\cref{fig: orthogonal}). 
Each pixel in AFM images transformed by pose $T$ and filtered by mask $M$ represents a sample ray.
The loss function is the disparity between the AFM data and the depth value derived from differentiable volume rendering along the ray. We then optimize the MLP network parameters $\theta$ through back propagation~\cite{backpropagation}. Moreover, we also use the multiresolution hash encoding technique~\cite{instant_ngp} to accelerate the training process.
Upon completing the training, we can query the SDF value of any spatial point by inferring the network. 
Based on the fact that the zero set of SDF represents the structure surface, the Marching Cubes algorithm~\cite{Marching_cubes_98} is finally utilized to extract the 3D surface model of the micro-/nanostructures (\cref{fig: pipeline}d).

\subsection{Reconstruction of Two-photon Lithography Structures}
\begin{figure*}[h]
    \centering
    \includegraphics[width=\linewidth]{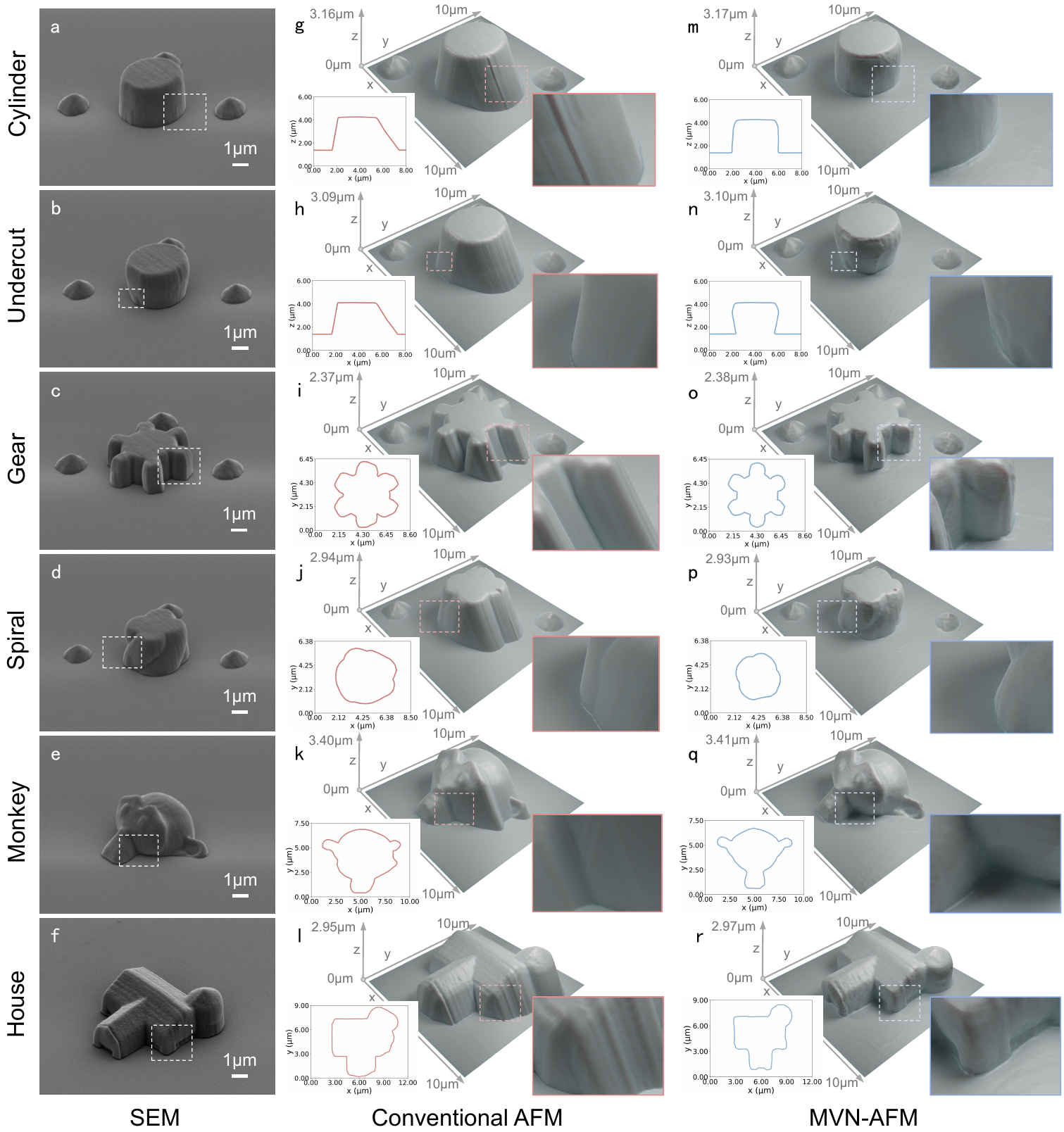}
    \vspace{-1.5em}
    \caption{ \textbf{MVN-AFM reconstructs the surface models of two-photon lithography microstructures.} \textbf{a-f} SEM photos of TPL microstructures. \textbf{g-l} 3D models of TPL microstructures' conventional AFM scanning data. 
    \textbf{m-r} 3D models of TPL microstructures reconstructed by MVN-AFM. 
    \textbf{g, h, m, n} Include the cross-section profiles in the x-z plane.
    \textbf{i-l, o-r} Include the cross-section profiles in the x-y plane.
    More visualizations can be found in Supplementary \cref{fig: TPL_SEM_top_down} and Supplementary Movie \textcolor{red}{1}.
}
    \label{fig: TPL}
    \vspace{-1em}
\end{figure*}

In this section, we evaluate the proposed MVN-AFM on microstructures printed by TPL technology.
The TPL technology, which focuses a femtosecond laser into tiny voxels in a photosensitive resist, enables 3D printing of a given Computer-Aided Design (CAD) model with sub-100 nm resolution through the two-photon polymerization (TPP) process~\cite{TPL_100nm}.
To fully demonstrate the performance of MVN-AFM on complex 3D microstructures, we printed a set of samples with different geometrical features. Specifically, we printed six structures (Supplementary \cref{fig: sim_models}): cylinder, undercut, spiral, gear, monkey, and house. 
For centrosymmetric structures, we incorporated three small cones around each microstructure to indicate their orientation, as depicted in the first four rows of \cref{fig: TPL}. This step is unnecessary for non-centrosymmetric structures, such as the monkey and the house. The height of these microstructures varies between 2 {\textmu}m and 3.5 {\textmu}m. We performed AFM scans in tapping mode, with a scan size of 10 {\textmu}m $\times$ 10 {\textmu}m and 256 lines of 256 points for each AFM image.

The cylinder (\cref{fig: TPL}a) is a representative structure that challenges conventional AFM scanning~\cite{11degree} and previous tilting methods. Unlike grating structures, the vertical annular sidewall cannot be divided into distinct left and right sections. 
Next, the undercut (\cref{fig: TPL}b) is a prevalent structural feature in semiconductor manufacturing \cite{undercut0}. 
This structure differs from the cylinder by having a sloped sidewall.
We further constructed the gear (\cref{fig: TPL}c), a mechanical structure frequently encountered in Micro-Electro-Mechanical systems (MEMS)\cite{MEMS_gear}.
The spiral (\cref{fig: TPL}d) is distinguished by an intricate array of rotating curved concave and convex structures on its sidewall.
Furthermore, we also conducted tests using the Suzanne Monkey (\cref{fig: TPL}e), a standard model in computer graphics ~\cite{Blender}.
Unlike the previous columnar structures, this model poses unique challenges due to its curved features and the indistinct boundary between its top surface and sidewalls. 
We finally designed a house structure (\cref{fig: TPL}f) that included shapes with both planar and curved features, along with detailed elements like grooves on the sidewalls to represent doors and windows.

\begin{figure*}[h]
    \centering
    \includegraphics[width=\linewidth]{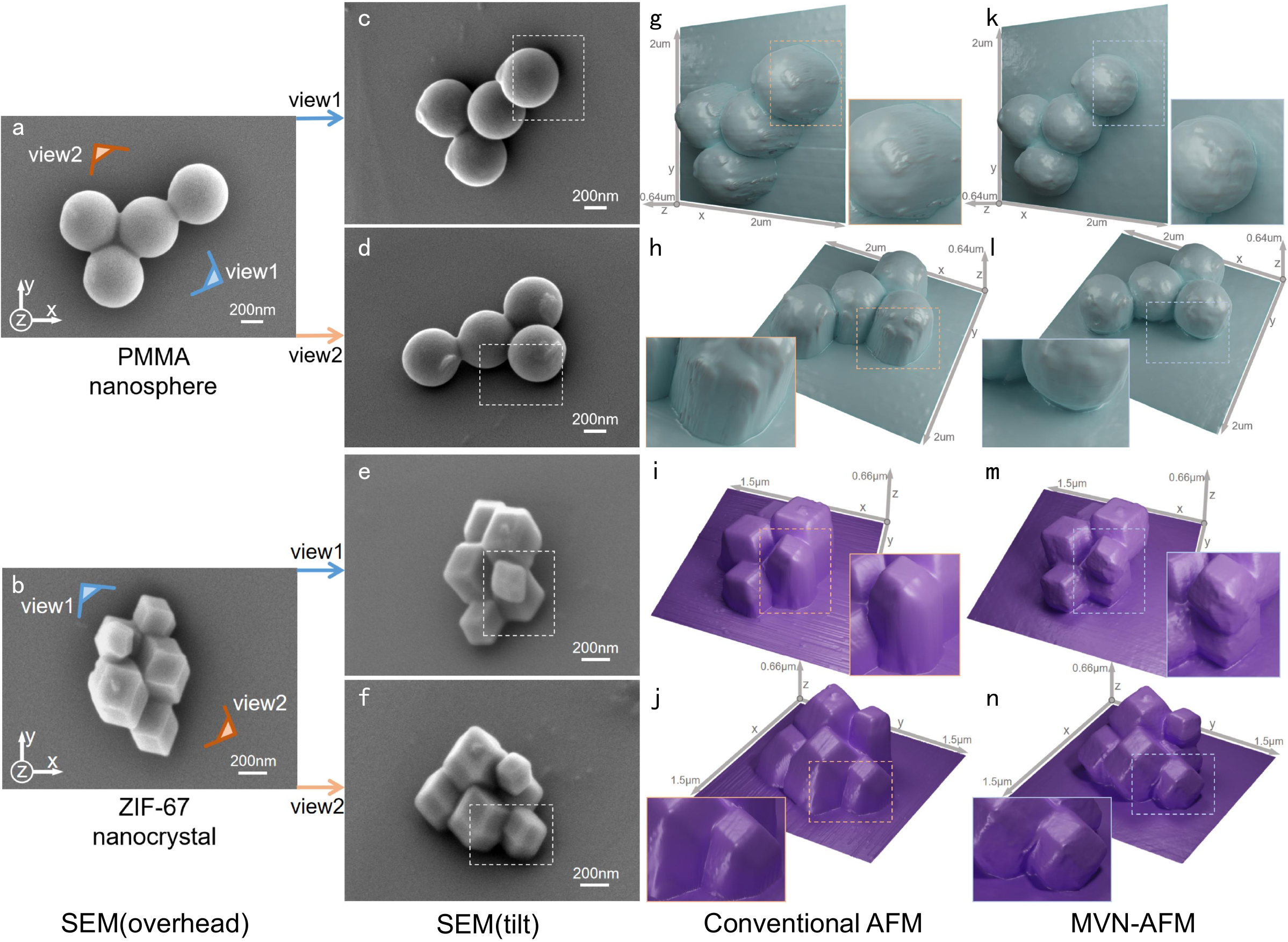}
    \vspace{-1em}
    \caption{ \textbf{MVN-AFM reconstructs the surface models of nanoparticles.} \textbf{a, b} SEM photos of nanoparticles in the overhead view.  \textbf{c-f} SEM photos of nanoparticles in the tilt view. \textbf{g-j} 3D models of nanoparticles' conventional AFM scanning data. \textbf{k-n} 3D models of nanoparticles reconstructed by MVN-AFM. \textbf{i, m} The local zoom reveals two stacked-up nanocrystals. 
    More visualizations can be found in Supplementary Movie \textcolor{red}{2}.
    }
    \label{fig: particle}
    \vspace{-1em}
\end{figure*}

In conventional AFM scanning, the results are a mixture of incomplete surface geometry and artifacts, which do not accurately represent the sample surface. As illustrated in \cref{fig: TPL}g and h, despite the vast difference in sidewall geometry, the scanning result of the undercut is indistinguishable from that of the cylinder model. Some detailed features are also virtually invisible, such as the doors and windows in the house model (\cref{fig: TPL}l). The cross-sectional profiles reveal significant distortion of these scanning results, which may lead researchers to misjudge the actual shape of these samples. 
Moreover, it is obvious that manually separating artifacts from the AFM scans of these intricate structures is almost impractical.

In contrast, the proposed MVN-AFM framework effectively eliminates artifacts while precisely merging geometric information from multi-view AFM scanning into accurate and comprehensive 3D models. 
These reconstructed models align consistently with SEM photos and demonstrate the surface of these samples. These models clearly differentiate between the cylinder (\cref{fig: TPL}m) and undercut (\cref{fig: TPL}n) structure, precisely reconstruct the gear's teeth (\cref{fig: TPL}o), and capture the correct orientation of the spiral threads(\cref{fig: TPL}p) and the monkey's subtly inward-curving side faces (\cref{fig: TPL}q). Even the minutely detailed grooves (\cref{fig: TPL}r) on the house sidewalls are observable.

\subsection{Reconstruction of Nanoparticles}

To further demonstrate the generalization of MVN-AFM on structures with smaller sizes and different geometry features, we selected some widely used nanoparticles, including PMMA nanospheres and ZIF-67 nanocrystals.

The spherical~\cite{polymeric_nanoparticles} is a typical shape of nanoparticles with extensive applications.
The characteristics of nanospheres depend significantly on their size and surface structure~\cite{Nanomedicines}, making accurate 3D reconstruction valuable for their research.
To evaluate the effectiveness of our proposed method on spherical structure, we chose PMMA~\cite{PMMA_drug, PMMA1, PMMA2} with a diameter of about 500 nm, a widely used type of polymeric nanosphere. 
In \cref{fig: particle}g and h, it is evident that the artifacts in the conventional AFM scanning data are seamlessly connected with the top curved surface of the nanospheres, and the overall shape does not exhibit a spherical appearance. 
Under these circumstances, manually distinguishing artifact boundaries in AFM scanning as in previous methods becomes unachievable, and the details on the sides of the nanospheres are entirely lost, posing a challenge for researchers to accurately determine the size and structure of these nanospheres.
In contrast, MVN-AFM demonstrates its advanced capabilities by accurately reconstructing several adherent nanospheres, each mirroring the shape observed in SEM photographs (\cref{fig: particle}c, d). Importantly, the tilt scanning feature of MVN-AFM captures the curved surface information on the sides of nanospheres. This information is seamlessly integrated, resulting in complete, artifact-free spherical reconstructions (\cref{fig: particle}k, l).

\begin{figure*}[h!]
    \centering
    \includegraphics[width=1.0\linewidth]{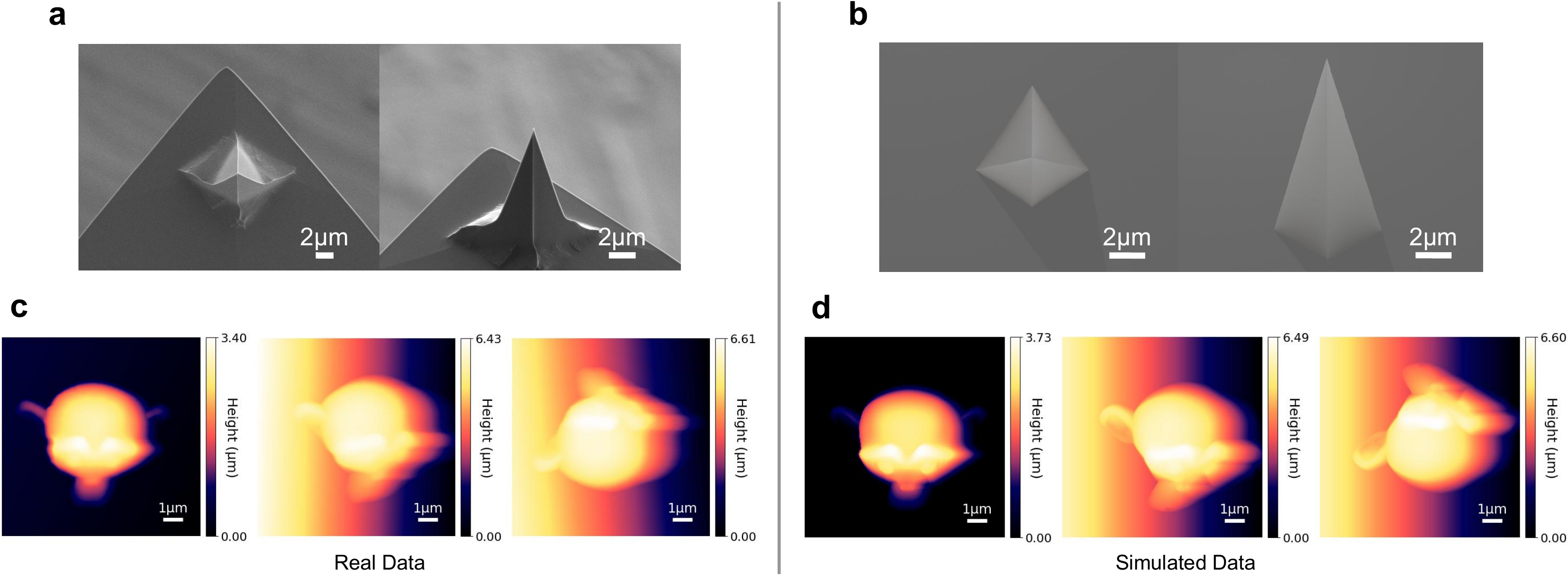}
    \vspace{-2em}
    \caption{ \textbf{The consistency between our simulated AFM data and the real experiment data.} \textbf{a} SEM photos of the real AFM probe used in the TPL experiment. \textbf{b} The probe model in the simulation environment. \textbf{c} The real AFM images captured by the multi-view tilt scanning of the monkey structure in the TPL experiment. \textbf{d} Simulated multi-view AFM images constructed by simulating the collision between the probe and the structure surface. 
 }
    \label{fig: sim_data}
\end{figure*}

\begin{figure*}[h!]
    \centering
    \includegraphics[width=0.7\linewidth]{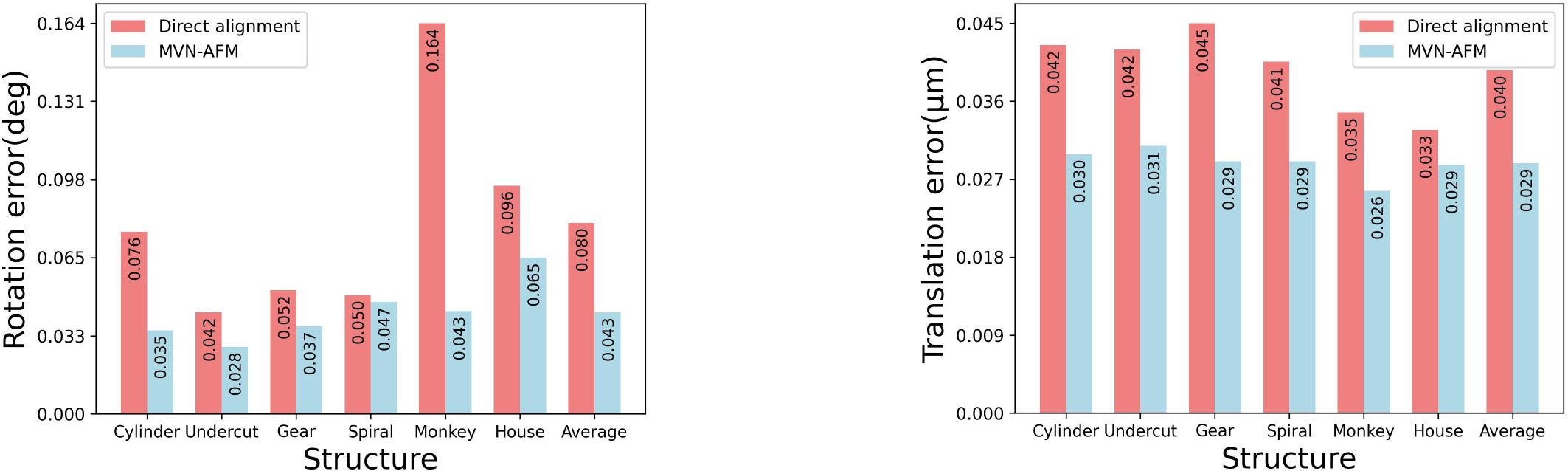}
    \vspace{-1em}
    \caption{ \textbf{Evaluation of MVN-AFM's improvement in multi-view data alignment accuracy.} Comparison of the absolute pose error, that is, rotation and translation error (lower is better), of direct alignment of raw AFM data and the alignment method of MVN-AFM.
 }
    \label{fig: pose_error}
    \vspace{-1em}
\end{figure*}

Next, we selected ZIF-67, a cubic symmetric nanocrystal, as a representative crystal-like nanoparticle to assess the effectiveness of our method. 
ZIF-67~\cite{ZIF-67_application} and its derivatives exhibit various excellent properties, leading to their extensive attention and research~\cite{synthesis_ZIF67, ZIF67_Electrochemically, ZIF67_Catalyst}. 
The morphological characteristics and size of nanocrystals can be tailored by manipulating experimental conditions during synthesis, leading to variations in their properties~\cite{MOF_ZIF-67, ZIF_size, metal_nanocrystals}. Therefore, obtaining accurate 3D surface models of nanocrystals is of paramount importance.
In the SEM photos (\cref{fig: particle}e, f), the ZIF-67 nanocrystals exhibit a distinct polyhedral shape, ranging in size from about 100 nm to 500 nm.
However, conventional AFM results (\cref{fig: particle}i, j) only partially demonstrate the top surface of the crystals, resulting in an overall blurred representation of the particles' shape.
Furthermore, in scenarios where multiple crystals aggregate, as illustrated in our example, only the uppermost crystal in the stack is visible in the conventional AFM scanning (\cref{fig: particle}i), with the underlying crystal completely obscured by the top crystal and associated probe artifacts. 
On the contrary, the surface model reconstructed by our method (\cref{fig: particle}m, n) accurately captures the polyhedral shape of the ZIF-67 crystals, delineating their side planes and edges with precision. Even in cases where the particles are stacked up, the MVN-AFM method successfully reveals the bottom crystal (\cref{fig: particle}m), typically obscured in conventional scans, and accurately represents the arrangement of the particles in the stack, aligning with the SEM photograph (\cref{fig: particle}e). 

In our nanoparticle experiments, PMMA nanospheres and ZIF-67 nanocrystals differ significantly from the previous TPL microstructures in terms of material compositions, geometric features, and particle sizes.
MVN-AFM precisely reconstructs these diverse samples by the exact same procedure and parameters, showcasing its outstanding generalizability and potential applicability in a broad spectrum of micro-/nanostructure research.

\subsection{Evaluation on Simulated Data}

\begin{figure*}[h!]
    \centering
    \includegraphics[width=\linewidth]{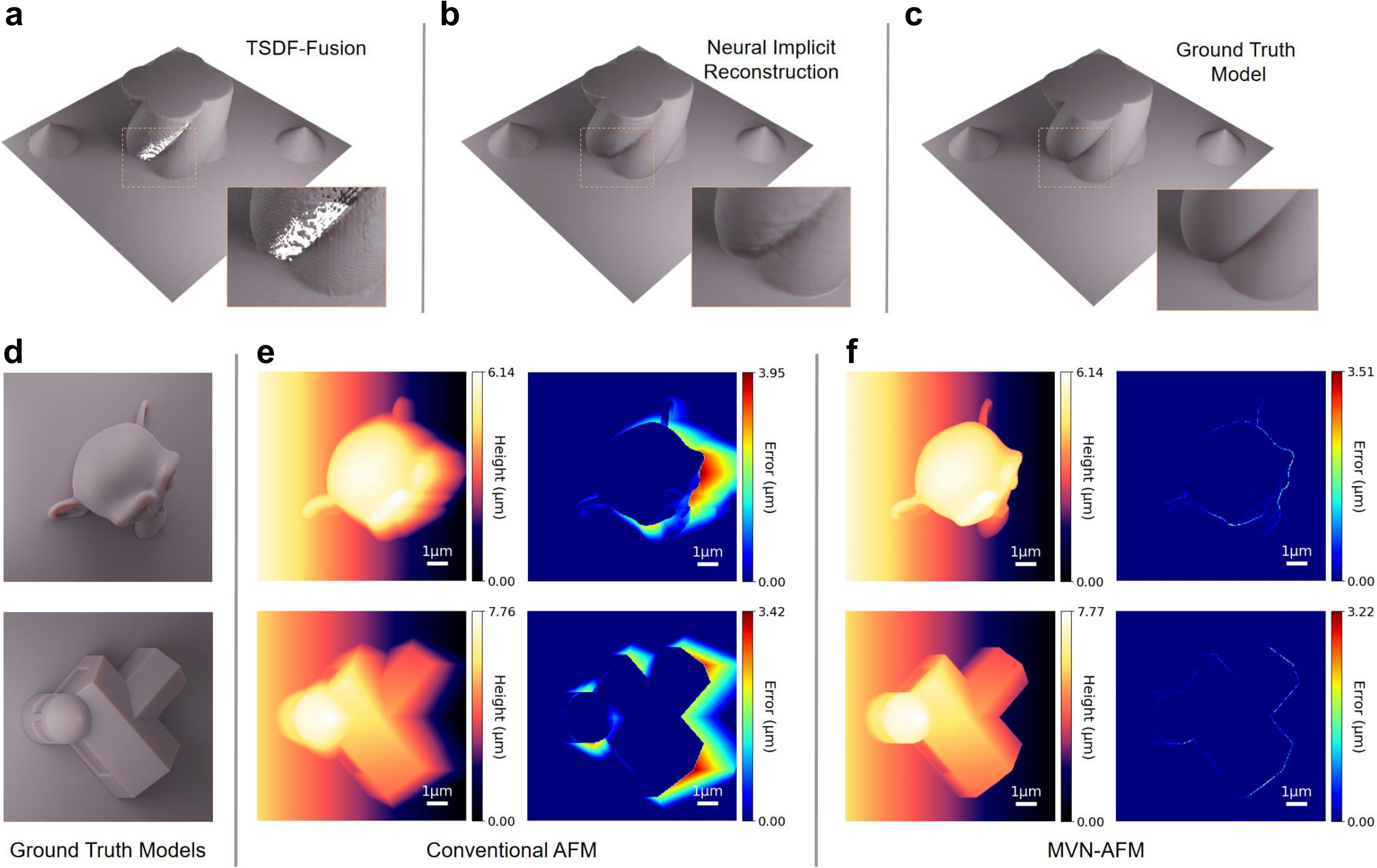}
    \vspace{-1.5em}
    \caption{ \textbf{Evaluation of MVN-AFM’s improvement in reconstructed 3D models.}
    \textbf{a} Given a set of masked and posed multi-view AFM images, the spiral model reconstructed by the TSDF Fusion method. The local zoom shows the voids on the reconstructed model.
    \textbf{b} Given the same set of masked and posed multi-view AFM images, the spiral model reconstructed by the neural implicit surface reconstruction. 
    \textbf{c} The ground truth model of the spiral.
    \textbf{d} The images of the monkey and house models in the given viewpoints.
    \textbf{e} The simulated AFM images and corresponding height error maps.
    \textbf{f} The topography images of 3D models reconstructed by MVN-AFM and corresponding height error maps. More visualizations can be found in Supplementary \cref{fig: more_error_map}
 }
    \label{fig: sim_result}
\end{figure*}

\begin{table*}
  \centering
  \small
  \begin{tabular}{c c c c c c c c c}
    \toprule
    MAE ({\textmu}m) &Cylinder&Undercut&Gear&Spiral&Monkey&House&Average\\
    \midrule
    Conventional AFM & 0.2587 & 0.2642 & 0.1750 & 0.2456 & 0.2043 & 0.2414 & 0.2315\\
    MVN-AFM & 0.0131 & 0.0137 & 0.0204 & 0.0128 & 0.0122 & 0.0173 & 0.0149\\
    \bottomrule
  \end{tabular}
  \vspace{-1em}
  \caption{ \textbf{The error comparison of conventional AFM images and MVN-AFM.} The mean absolute error (lower is better) of the input conventional AFM images and the topography images from 3D models reconstructed by MVN-AFM.}
  \vspace{-1em}
  \label{tab:depth_error}
\end{table*}

To complement the previous qualitative comparisons on real experimental data, we embarked on quantitative evaluations using a set of simulated AFM data. We generated these data based on the CAD models of structures in the TPL experiment. The simulation environment allows for the precise determination of the spatial relationships between multi-view AFM data and access to an accurate surface model of the sample, a feat challenging to achieve in real-world experiments.
To ensure that the simulated data closely mimics real AFM scanning conditions, we developed a simulated probe model. This model is based on the quadrilateral pyramid probe (\cref{fig: sim_data}a) utilized in our TPL experiments. Considering the nanoscale curvature of the actual AFM probe is negligible compared to the microscale dimensions of the TPL samples, we simplified the probe representation into a pyramid shape (\cref{fig: sim_data}b).
The simulation of AFM scanning was then carried out by modeling the rigid body collision~\cite{BioAFMviewer} between the probe and the sample models. 
As depicted in \cref{fig: sim_data}c and d, the simulated data exhibit a high degree of similarity to the real AFM data in terms of the overall shape and the presence of artifacts.

First, we focus on showcasing the enhancements MVN-AFM brings to the alignment accuracy of multi-view AFM data.
We quantified the error in this alignment process by comparing the rotation component $R$ and the translation component $t$ of pose $T$ with the accurate $\overline{R}$ and $\overline{t}$ for each viewpoint acquired in the simulation environment.
As illustrated in \cref{fig: pose_error}, we present a comparative analysis between the alignment method in MVN-AFM and the direct ICP alignment of raw AFM data, which includes artifacts.
The analysis reveals that MVN-AFM achieves a substantial improvement in alignment accuracy, evidenced by an impressive average reduction of 46\% in rotation errors and 27\% in translation errors. 
These results not only demonstrate the negative impact of artifacts present in AFM data on the precision of data alignment but also highlight the efficacy of MVN-AFM in mitigating these challenges.

In the subsequent analysis, we compare the models reconstructed by two prominent multi-view depth fusion techniques: the neural implicit method and TSDF (Truncated Signed Distance Function) Fusion~\cite{Kinectfusion}.
Our method interprets AFM images as depth images from virtual orthogonal cameras, framing the challenge as the depth fusion problem in computer vision.
Depth fusion techniques are categorized into traditional~\cite{Kinectfusion, classical_recon1, classical_recon2} and neural implicit methods~\cite{Neural_survey, neuralfusion,bvn-fusion,imap,Neus}. 
The TSDF Fusion is a widely used traditional method that efficiently fuses multi-view depth data by dividing the 3D space into weighted discrete voxels and updating these weights according to the depth information along the pixel ray.
However, multi-view AFM scanning of micro-/nanostructures presents unique challenges, particularly the uneven sampling density (Supplementary \cref{fig: uneven_density}a) due to restricted tilt angles and limited viewpoints during the scanning process.
This limitation often leads to regions with sparse sampling, such as the sidewall grooves of the spiral model (\cref{fig: sim_result}c).
In the context of TSDF Fusion (Supplementary \cref{fig: uneven_density}b), unintersected voxels in sparsely sampled regions demonstrate as voids in the reconstructed model, a limitation evident in \cref{fig: sim_result}a. 
Conversely, the neural implicit method, which represents the 3D model as a continuous neural network,  exhibits a remarkable ability to construct a smooth and complete surface model, even with limited sample points, as depicted in \cref{fig: sim_result}b. 
This capability of the neural implicit method to effectively handle sparse data and reconstruct intricate surfaces makes it more suitable for the 3D reconstruction of multi-view AFM data in the MVN-AFM framework.

Finally, we evaluated the accuracy of the topography in the 3D surface models reconstructed by MVN-AFM. 
Our simulation environment enables the capture of precise surface topography unaffected by the probe's shape.
The difference between accurate surface topography and the AFM images reveals substantial artifacts (\cref{fig: sim_result}e), particularly around the edges and at the sharper geometric features of the structure in the raw AFM data. These results underscore the complexity of artifacts in AFM images of intricate structures and highlight the challenges associated with their manual removal.
The visualization of the difference between the topography images from 3D models of MVN-AFM and the accurate topography images (\cref{fig: sim_result}f) clearly indicates that MVN-AFM is highly effective in eliminating the artifacts present in the AFM data. Moreover, it successfully integrates accurate surface geometric information from various viewpoints, significantly diminishing the surface topography error.
To quantify these improvements, we calculated the average of the absolute pixel error values across multiple viewpoints for each model. As summarized in \cref{tab:depth_error}, this analysis reveals that MVN-AFM achieved an exceptional average reduction of 94\% in topography error for each structure, affirming the high accuracy of the 3D models reconstructed by MVN-AFM.

\section{Discussion}
In this work, we introduce MVN-AFM, a framework for 3D surface reconstruction of intricate micro-/nanostructures using multi-view AFM scanning data.
We propose a novel iterative optimization method to simultaneously align the multi-view data and remove artifacts in the AFM image, achieving higher alignment accuracy.
To the best of our knowledge, we are the first to utilize the neural implicit surface reconstruction technique in the field of nanotechnology, which enables fusing spatially overlapping multi-view AFM data into an accurate 3D model.
MVN-AFM shows considerable practical value.
Extensive experiments demonstrate the superior capability of MVN-AFM on diverse micro-/nanostructures, including microstructures printed by TPL, PMMA nanospheres, and ZIF-67 nanocrystals.
The 3D models reconstructed by MVN-AFM provide researchers with a more comprehensive representation of micro-/nanostructures than what is achievable with conventional AFM scanning and 2D SEM images.
The success of MVN-AFM across these varied samples, each with distinct geometries, types, and sizes, robustly affirms its effectiveness and broad applicability in nanofabrication, nanoparticles, and many other fields.
Importantly, MVN-AFM only requires a conventional AFM system and a standard AFM probe to achieve these results.
This aspect makes MVN-AFM a more accessible and cost-effective option for researchers to analyze intricate 3D micro-/nanostructures.

Our framework is efficient and flexible. 
While multi-view AFM data provides more surface information, it also increases the time for AFM scanning.
We tested the effect of reconstruction using different numbers of tilted AFM data(Supplementary \cref{fig: tilt_number} and Supplementary \cref{tab:tilt_view_error}). We found that for the structures in the TPL experiment, the reconstruction quality converged with only eight tilt scans.
The scanning time for one AFM image is about 4.5 minutes, and considering the time required to switch to different scanning directions, multi-view scanning takes about 2 hours for a single structure.
Next, given a set of multi-view AFM data, our framework takes about 10 minutes to complete the 3D reconstruction.
Notably, the basis of our algorithm is the multi-view consistency of the accurate surface topography and the multi-view inconsistency of image artifacts in multi-view AFM data. Because our algorithm does not take parameters such as scanning number, tilt angle, and probe shape as prior information, users have the flexibility to adjust these parameters according to their requirements.

Here, our study underscores the significant potential of integrating nanotechnology with neural implicit representations~\cite{implicit_neural, img_neural, d-nerf}, an emerging and rapidly evolving field in computer vision.
Specifically, we employ neural implicit surface reconstruction methods, where a neural network effectively represents a continuous SDF in space.
Because of the continuous nature of neural networks, it is more suitable for representing geometric surface models that are inherently continuous in space than traditional discrete methods, as demonstrated in numerous recent works~\cite{Neural_survey, neuralfusion, bvn-fusion}.
Our research further reveals the successful application of this technique in the reconstruction of 3D micro-/nanostructures with multi-view AFM data.

Our methodology's foundational assumption is that the sample remains static during the multi-view AFM scanning process because the multi-view AFM data alignment step depends on the consistency of geometric features across different views. Therefore, our method is unsuitable for dynamic samples, such as living cells or samples prone to deformation during scanning. 
Precisely reconstructing the deformation process of nanostructures is widely demanded in many research, which points to a promising direction for future work. 
One possible solution is applying our method to High-Speed Atomic Force Microscopy (HS-AFM)~\cite{HS-AFM}, which allows observing the dynamic action of nano-objects.

\section{Methods}

\subsection{Hardware and Software Requirments of MVN-AFM}
In our experiments, all the code of MVN-AFM was run on a computer with an Intel i9-13900KF CPU, an Nvidia RTX4090 GPU, 64 gigabytes of RAM, and a Linux operation system with a 5.15.0 kernel version, which is a typical configuration of the current lab workstation computer. In order to run the code of MVN-AFM properly, it requires at least one graphics card with memory larger than 12 gigabytes. We use Open3D~\cite{Open3d} 0.17.0, an open-source Python library, to handle the 3D data.
Our implementation of neural implicit surface reconstruction is based on an open-source repository~\cite{instant-nsr-pl} of hash encoding~\cite{instant_ngp} and NeuS~\cite{Neus}, and the network is built on the deep learning framework PyTorch~\cite{pytorch} 1.13.1. 

 \subsection{Multi-View AFM Scanning}

All multi-view AFM images in our experiments were acquired through a commercial AFM (Dimension ICON, Bruker). The AFM probe we used was the TESPA-V2 (Bruker), which has a height of 15 {\textmu}m, an overall shape of a quadrilateral pyramid, a front angle of 25$^{\circ}$, a back angle of 17.5$^{\circ}$, and a side angle of 20$^{\circ}$.
The stage has a 24 mm$\times$24 mm square bottom, and the height is 16 mm with a 30$^{\circ}$ tilt angle. The turntable can hold a 4 mm$\times$4 mm sample. The whole stage can be placed directly into a commercial AFM and does not collide with any part of the AFM during scanning.
In our experiments, we rotate 45$^{\circ}$ each time between two adjacent scans and obtain eight tilt scans around the sample. Together with a conventional overhead view, nine scans are acquired per sample.
For every view, we obtain an AFM image with 256 lines of 256 points by AFM working in tapping mode at a frequency of 1 Hz.

Precisely localizing the identical region across multi-view AFM scans is a critical step in the data capture process.
The methodologies for achieving this localization are diverse and can be tailored to the unique characteristics of the experiment sample.
In our TPL experiments, we utilized polymer grid markers printed around the sample to assist in localization by the optical microscope in the AFM system.
For experiments of nanoparticles, we constructed scored markers on mica bases. 
These are just examples of the various strategies that can be adopted for localization, with other available methods including the use of a Transmission Electron Microscopy (TEM) index grid ~\cite{TEM_locate, TEM_locate2} or the creation of noticeable artificial markers~\cite{mica_locate}.
The common destination of these techniques is to ensure that the specific structure for 3D reconstruction can be precisely and efficiently located within the AFM system.

\subsection{Data Alignment and Mask Solving}

First, we claim some basic concepts in this step.
Each AFM image is equivalent to a set of 3D points under an AFM coordinate system (Supplementary \cref{fig: coordinate}), with the z-axis being the position feedback direction of AFM and the x-y plane being the probe scanning plane. We define the AFM coordinate system of the overhead view image of the sample as the destinated sample coordinate system. 
Moreover, we define a corresponding virtual orthogonal depth camera for each AFM image (Supplementary \cref{fig: orthogonal}). 
With a given set of raw AFM data, we convert the AFM height information $h$ into a depth value $d$ for a virtual orthogonal camera parallel to the x-y plane, $d = \alpha - h$, where $\alpha$ is the assumed height of the camera. The value of $\alpha$ is simply ensured all $d$ to be positive.
Each AFM data point is treated as a ray $r = o + d\vec{v}$, originating from the pixel position $o$ on the imaging plane and extending along the direction $\vec{v}$ of the camera to the depth $d$.

In the initialization and the M-step, we applied the point-to-plane ICP algorithm~\cite{ICP} to align the data points filtered by mask $M$ of AFM images and get a set of transformations $T$. 
In the E-step, We compute the artifact mask $M$ of each AFM image by a cross-validation method. 
In detail, we first transform each camera ray $r$ to the sample coordinate system to obtain the ray $r'$ by the currently solved $T = \{ (R, t) \mid R \in \mathbb{R}^{3 \times 3}, \, t \in \mathbb{R}^3 \}$, where $r' = o' + d\vec{v}'$, $o' = Ro + t$, and $\vec{v}' = R\vec{v}$. 
Subsequently, we generate $n$ sets of meshes by connecting spatial points corresponding to neighboring pixels in $n$ AFM images within the sample coordinate system. 
Next, we compute the intersection of each ray with these meshes and obtain $n$ depth values, $D = \{d_1, d_2, ..., d_n\}$.
Due to a basic fact, the artifacts of tip-sample convolution cause an expansion of the overall topography~\cite{AFM_image_artifacts}, resulting in the height value of AFM scanning being larger than the actual sample height, equivalent to the smaller depth value. 
Therefore,  we consider a pixel as an artifact when $D_{\text{max}} - d > \phi$, where d is the measured depth of each pixel, and $\phi$ is set to 3\% of the AFM scan size initially and linearly reduced to 1\% with iteration, a value determined experimentally and applied consistently across all our experiments. The tiny threshold is set to make the algorithm robust to noises in the AFM data and inaccurate $T$ during the iteration process.
These iterative EM-steps reinforce each other. After a fixed number of iterations, five in our experiments, the resolved poses $T$, and masks $M$ are saved for subsequent steps.
For a set of multi-view AFM data, this process takes about 2 minutes.

\subsection{Neural Implicit Surface Reconstruction}

In the neural implicit surface reconstruction step, we train a multi-resolution hash table with learnable parameters and an MLP neural network named the SDF network by the aligned and masked multi-view AFM data (Supplementary \cref{fig: network_pipeline}). 
In the training process, we sample 3D points along the ray $r'$ of pixels filtered by mask $M$.
First, the 3D point coordinate is encoded by multiresolution hash technology~\cite{instant_ngp}. 
Here, we use 16 resolution levels, each obtaining a 2-dimensional feature vector.
Concatenating the hash encoding and the 3D coordinate, we get a 35-dimensional feature as input of the SDF network. 
The SDF network is a one-layer MLP network with 64 hidden sizes and ReLU activation, which maps the input feature to an SDF value at that 3D point.
The SDF value of each point is converted to a density value through the unbiased and occlusion-aware weight function proposed by NeuS~\cite{Neus}. Then, the density values of sample points along the ray are accumulated by the differentiable volume rendering method to obtain the depth value $\widehat{d}$ of that ray.  
The loss function $L$ consists of a depth error term $L_{\text{depth}}$ and a regularization term $L_{\text{reg}}$: 
\begin{equation}
L_{\text{depth}} = \frac{1}{b}\sum_{p}^{b} (\widehat{d_p} - d_p)^2,
\end{equation}
\begin{equation}
L_{\text{reg}} = \frac{1}{bm} \sum_{p,q}^{b,m} \left( \|
\mathbf{n}_{pq}\| - 1\right)^2.
\end{equation}
$L_{\text{depth}}$ is the mean square error (MSE) between the rendering depth value of each pixel and the AFM data supervision, where $b$ is the batch size and $p$ is the index. 
The regularization term~\cite{SDF_Loss} is used to constrain the SDF field represented by the network, where $\mathbf{n}$ is the normal of the sample point, $m$ is the number of sample points along a ray, and $q$ is the index. $L_{\text{reg}}$ facilitates a smooth and natural surface, commonly used in SDF-based neural implicit surface reconstruction methods. 
\begin{equation}
L = L_{\text{depth}} + \lambda L_{\text{reg}}.
\end{equation}
The weight $\lambda$ is 0.1 in our experiment.
During the network training, the Adam optimizer updates the network parameters with a learning rate of 0.001 to minimize the loss and perform 20,000 iterations. In one iteration, we randomly select 256 rays with 1024 sample points along the ray. 
Notably, Each set of network parameters can only represent a 3D model of one structure, so multi-view AFM images for different samples need to be trained from scratch.
The whole training time is about 8 minutes on an Nvidia RTX4090 GPU.
To visualize the 3D model, we divide the space into 256$\times$256$\times$256 voxels. Subsequently, the SDF values for each voxel are obtained through neural network inference, followed by the extraction of meshes using the Marching Cube algorithm~\cite{Marching_cubes_98}.
Unlike the traditional discrete voxel-based representation~\cite{Kinectfusion}, which requires the prior determination of a voxel division resolution, neural implicit surface representations do not have a resolution limitation. The network can infer SDF values at any location, enabling the generation of a mesh representation with arbitrary resolution.

\subsection{Constructing the Two-photon Lithography Structures}
We used a commercial photoresist IP-Dip2 (Nanoscribe GmbH) as our material. The IP-Dip2 was dropped on a glass substrate with a thickness of 170-190 {\textmu}m  (Borosilicate substrates, Nanoscribe GmbH) for fabricating the structures. We used A commercial Direct Laser Writing setup (Photonic Professional GT2, Nanoscribe GmbH) equipped with a 780 nm femtosecond laser (a repetition rate of 80 MHz, a pulse duration of 80-100 fs) and a 63$\times$, numerical aperture (NA) = 1.4 oil immersion objective to print the microstructures.  We imported the STL files into Describe 2.7 (Nanoscribe GmbH) to generate the executable job files.  We set the slice and hatching distances to 0.1 {\textmu}m for microstructures, the highest accuracy this machine can achieve.
These distances were set to 0.3 {\textmu}m for grid markers because they are only used for optical microscope localization, which has no high requirement for printing accuracy. 
The printing parameters were set to 30 mW of laser power and 10,000 {\textmu}m/s scanning speed. Then, we imported the executable job files to Nanowrite 1.8 (Nanoscribe GmbH) to start the job. After the printing process, the printed structures were developed with propylene glycol methyl ether acetate (PGMEA) for 20 minutes and isopropyl alcohol (IPA) for 5 minutes to wash out the unpolymerized resists at room temperature and leave the microstructures on the substrate.

\subsection{Constructing the Nanoparticle Samples}

PMMA nanosphere dispersion (500 nm) was purchased from the Jiangsu Zhichuan Technology Co., Ltd (China). ZIF-67 powder (300 nm) was purchased from the Nanjing Xianfeng Nano Co., Ltd (China). We used ethanol to dilute these nanoparticles, sonicated them for 10 minutes, and then deposited the suspension onto a 4 mm$\times$4 mm mica base. We performed multi-view localization of the particles of interest based on the markers of the mica surface around the region. 
We used the same view number, tilt angle, AFM scanning mode, and AFM probe as in the TPL experiment. 
The AFM scan size was 2 {\textmu}m$\times$2 {\textmu}m for PMMA nanospheres and 1.5 {\textmu}m$\times$1.5 {\textmu}m for ZIF-67 nanocrystals. We obtained SEM photos of these micro-/nanostructures by sputter-coating samples with platinum by sputtering apparatus (MCIOO, Hitachi) and then observing them with a field-emission scanning electron microscope (GeminiSEM 300, ZEISS).

\subsection{Constructing the Simulated Data}

The simulated data was generated using the 3D design software Blender~\cite{Blender} 3.3. Within Blender, we constructed models of the structures as well as the AFM probe. 
To mimic the conditions of our real-world multi-view AFM scanning, we set up orthogonal cameras within the software positioned to align with the scanning directions in our real experiment. 
Furthermore, to replicate the real-world experimental setup more accurately, we rotated the probe model by 11$^{\circ}$. This adjustment accounts for the inherent angle between the working cantilever of the AFM holder and the scanning plane in a real AFM system~\cite{11degree, AFM_image_artifacts}.
Next, we orthogonally projected the surface model onto these cameras and performed a convolution of the probe shape to generate simulated AFM images.
In the quantitative evaluation of the accuracy of solved poses, we extracted the precise poses of these cameras directly from Blender and evaluated the absolute pose error by an open-source tool EVO~\cite{evo_tool}.
We implemented the TSDF Fusion method based on an open source repository~\cite{tsdf-fusion-python} and added support for the orthogonal camera projection model that allows for the fusion of multi-view AFM data. We divided the space into 256$\times$256$\times$256 voxels to keep consistent with the setup of the mesh model extraction step in the neural implicit surface reconstruction.
When it comes to evaluating the topography images of the 3D models reconstructed by MVN-AFM, we employ the Mean Absolute Error (MAE) as our metric.
\begin{equation}
\text{MAE} = \frac{1}{mn} \sum_{i=1}^{m} \sum_{j=1}^{n} \left| \bar{h}_{ij} - h_{ij} \right|,
\end{equation}
where $n$ denotes the number of multi-view images, $m$ is the pixel number in each image, $\bar{h}$ is the accurate height value of a pixel, and $h$ denotes the value of pixels in raw AFM images or topography images from MVN-AFM.

\section{Data Availability}
The data that support the findings of this study are available from the corresponding
author upon reasonable request.

\section{Code Availability}
The source code of MVN-AFM is available at \href{https://github.com/zju3dv/MVN-AFM}{https://github.com/zju3dv/MVN-AFM}.

{\small
\bibliographystyle{naturemag}
\bibliography{egbib}
}

\section{Acknowledgements}
We thank A. Ren, L. Ma, and C. Wu for assistance in the two-photon lithography experiment. We are grateful to Y. Wang, J. Tang, and M. Duan for helpful discussions. We also thank the staff of the Analysis Center of Agrobiology and Environmental Sciences, Zhejiang University, for their support in SEM imaging. This work was partially supported by the National Natural Science Foundation of China (No.61932003 received by G.Z., No.51975522 and No.U22A20207 received by Y.-L.C.).

\section{Author Contributions}
S.C. conceived the idea and proposed this project; S.C. and M.P. performed experiments; S.C. wrote code and processed data; S.C. and Y.L. wrote the draft of the manuscript, and all co-authors proofread and revised the manuscript; G.Z., Y.-L.C., H.B., and B.-F.J. provided valuable suggestions including the experiment design and writing. G.Z. and Y.-L.C. supervised this project, including the framework design and improvement.

\section{Competing Interests}
The authors declare no competing interests.

\onecolumn{
    \centering
    \Large
    \vspace{1.0em}
    \textbf{Multi-View Neural 3D Reconstruction of Micro-/Nanostructures \\with Atomic Force Microscopy
    } \\
    \vspace{0.5em}Supplementary Information \\
    \vspace{1.0em}
}

\setcounter{table}{0}
\setcounter{figure}{0}
\setcounter{equation}{0}

\begin{figure*}[h]
    \centering
    \includegraphics[width=\linewidth]{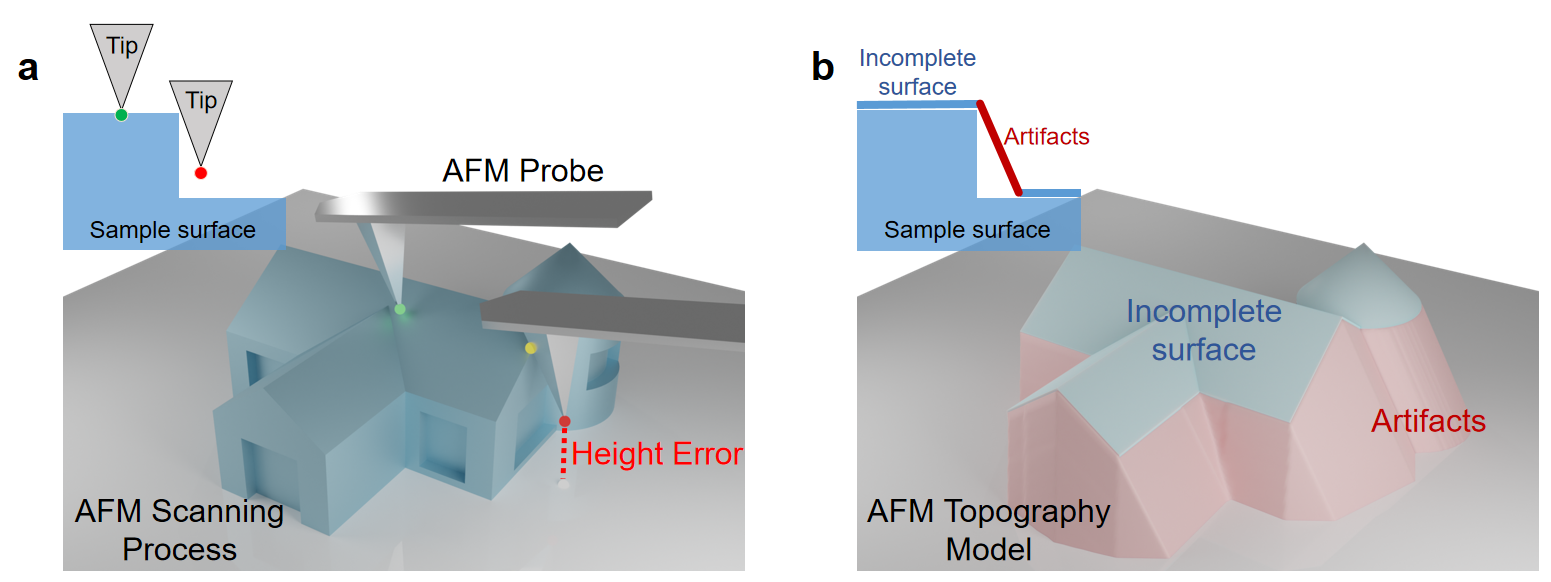}
    \caption[The limitations of conventional AFM scanning.]{ 
    \textbf{The limitations of conventional AFM scanning.} 
    \textbf{a} Illustration of conventional AFM scanning process. 
    AFM obtains topography information in the vertical direction by the interaction between the probe and the sample. However, when the side of the probe, rather than the probe's tip, touches the sample, AFM cannot get accurate height information. This phenomenon, which leads to artifacts in the AFM images, is called tip-sample convolution.
    \textbf{b} The 3D model of the conventional AFM scanning result. The scanning result is the combination of an incomplete structure surface (the blue part) and artifacts (the red part).
}
    \label{fig: afm_limit}
\end{figure*}

\begin{figure*}[h]
    \centering
    \vspace{2cm}
    \includegraphics[width=\linewidth]{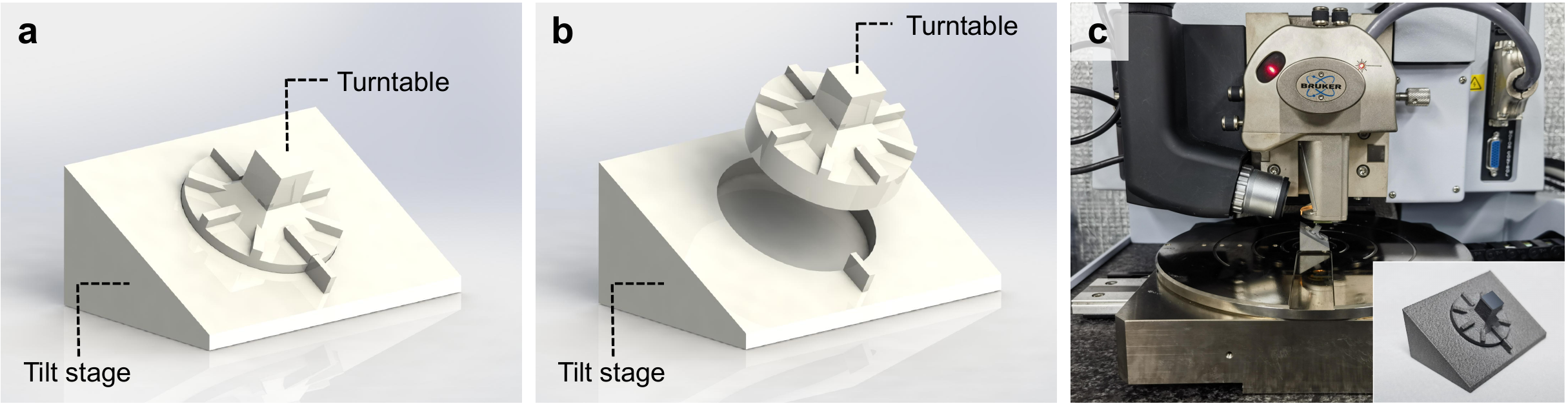}
    \caption[The tilt stage for multi-view AFM scanning.]{ 
    \textbf{The tilt stage for multi-view AFM scanning.} 
    \textbf{a, b} The design model of the tilt stage and the turntable in its center. 
    \textbf{c} The photos of the tilt stage and the AFM multi-view scanning process.
    }
    \label{fig: rotate_base}
\end{figure*}

\begin{figure*}[h]
    \centering
    \includegraphics[width=\linewidth]{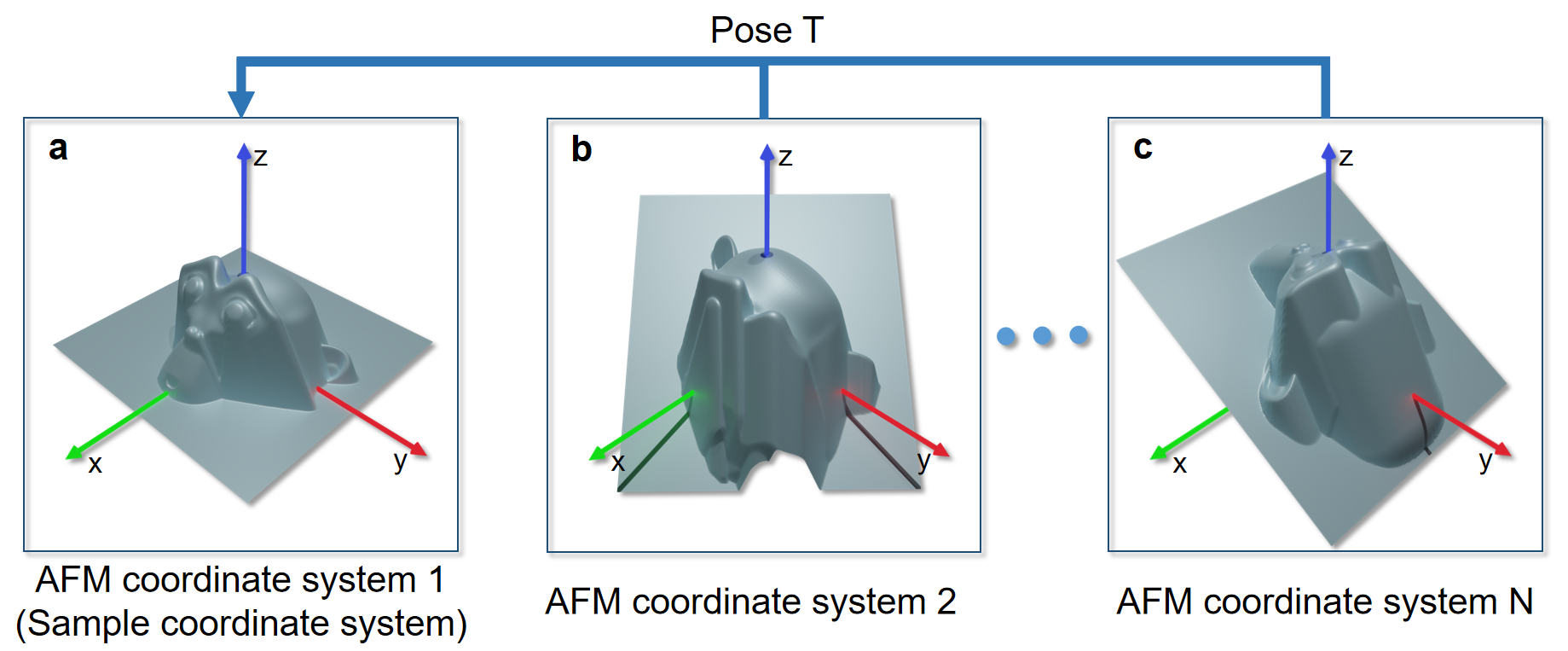}
    \vspace{-1.5em}
    \caption[The definition of coordinate systems.]{ 
    \textbf{ The definition of coordinate systems. }
    \textbf{a} The simulated overhead AFM scanning result. Its coordinate system is defined as the sample coordinate system.
    \textbf{b, c} The simulated multi-view tilt AFM scanning results. The pose $T$ transforms these data from multiple AFM coordinate systems to the sample coordinate system.
    }
    \label{fig: coordinate}
\end{figure*}

\begin{figure*}[h]
    \centering
    \includegraphics[width=0.6\linewidth]{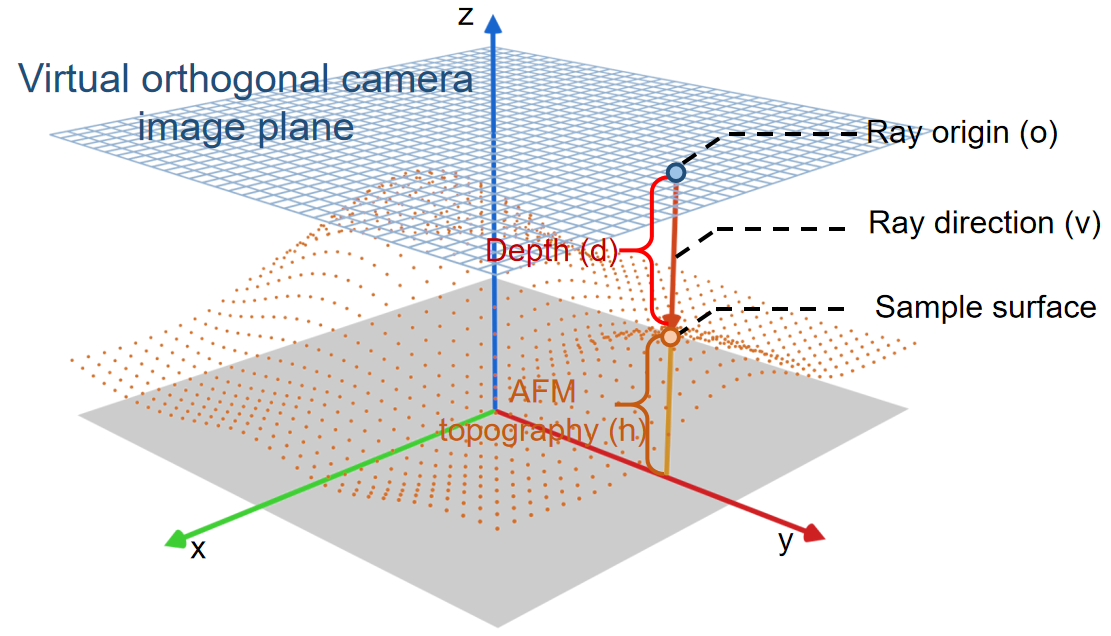}
    \caption[The definition of the virtual orthogonal camera model.]{ 
    \textbf{The definition of the virtual orthogonal camera model.}
    The raw AFM topography images are transformed into depth images of virtual orthogonal cameras in MVN-AFM. Each pixel of data with a height value (h) in the AFM image is considered a ray that starts at the origin (o), moves towards the direction (v), and terminates at a depth (d). 
    }
    \label{fig: orthogonal}
\end{figure*}

\begin{figure*}[h]
    \centering
    \includegraphics[width=0.9\linewidth]{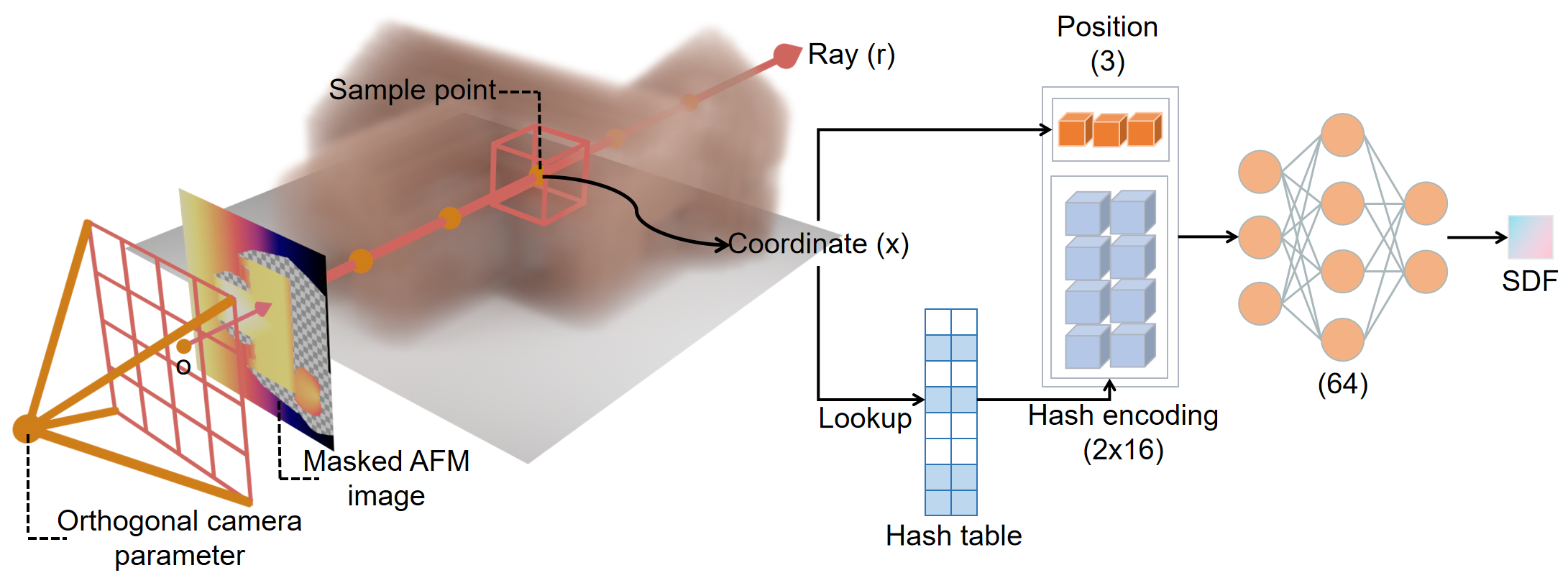}
    \caption[The pipeline of neural implicit surface reconstruction.]{ 
    \textbf{ The pipeline of neural implicit surface reconstruction. }
    The values under Position and Hash encoding show the sizes of the feature. The value under the network is the size of the hidden layer. 
    }
    \label{fig: network_pipeline}
\end{figure*}

\begin{figure*}[h]
    \centering
    \includegraphics[width=\linewidth]{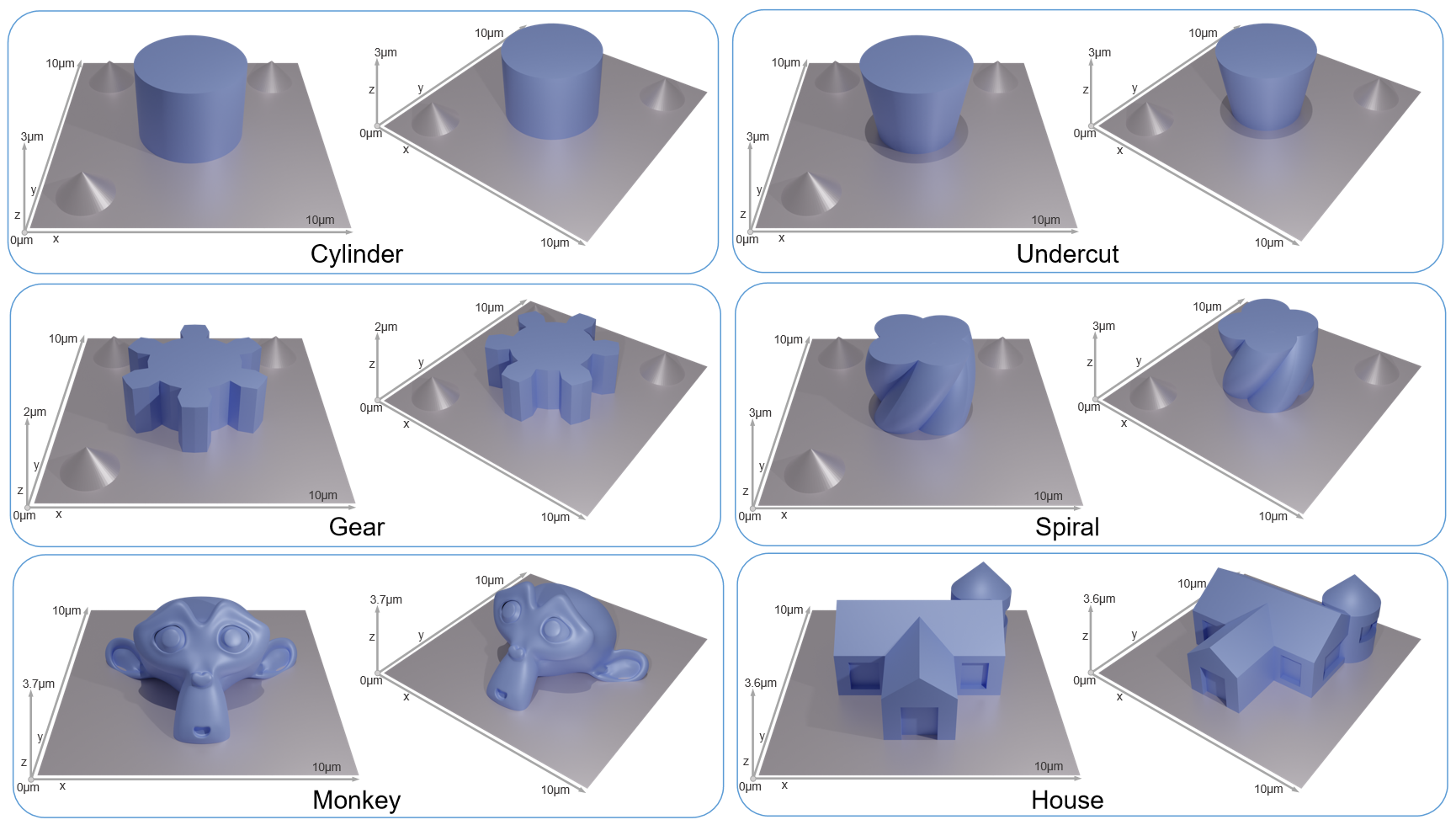}
    \vspace{-1.5em}
    \caption[Design models of TPL microstructures and simulated data.]{ 
    \textbf{Design models of TPL microstructures and simulated data.}
    Marked on the z-axis is the height of each microstructure. The size of the bottom plane of each microstructure is 10 {\textmu}m$\times$10 {\textmu}m.
    }
    \label{fig: sim_models}
\end{figure*}

\begin{figure*}[h]
    \centering
    \includegraphics[width=0.7\linewidth]{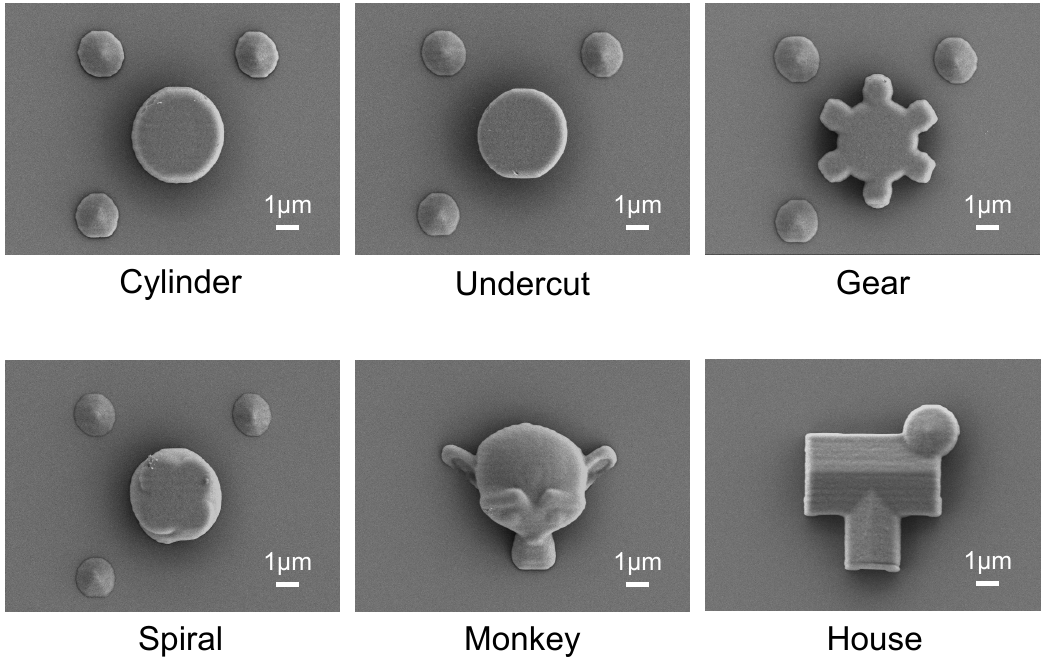}
    \caption[The SEM overhead view of TPL microstructures.]{ 
    \textbf{The SEM overhead view of TPL microstructures. }
    }
    \label{fig: TPL_SEM_top_down}
\end{figure*}

\begin{figure*}[h]
    \centering
    \includegraphics[width=0.95\linewidth]{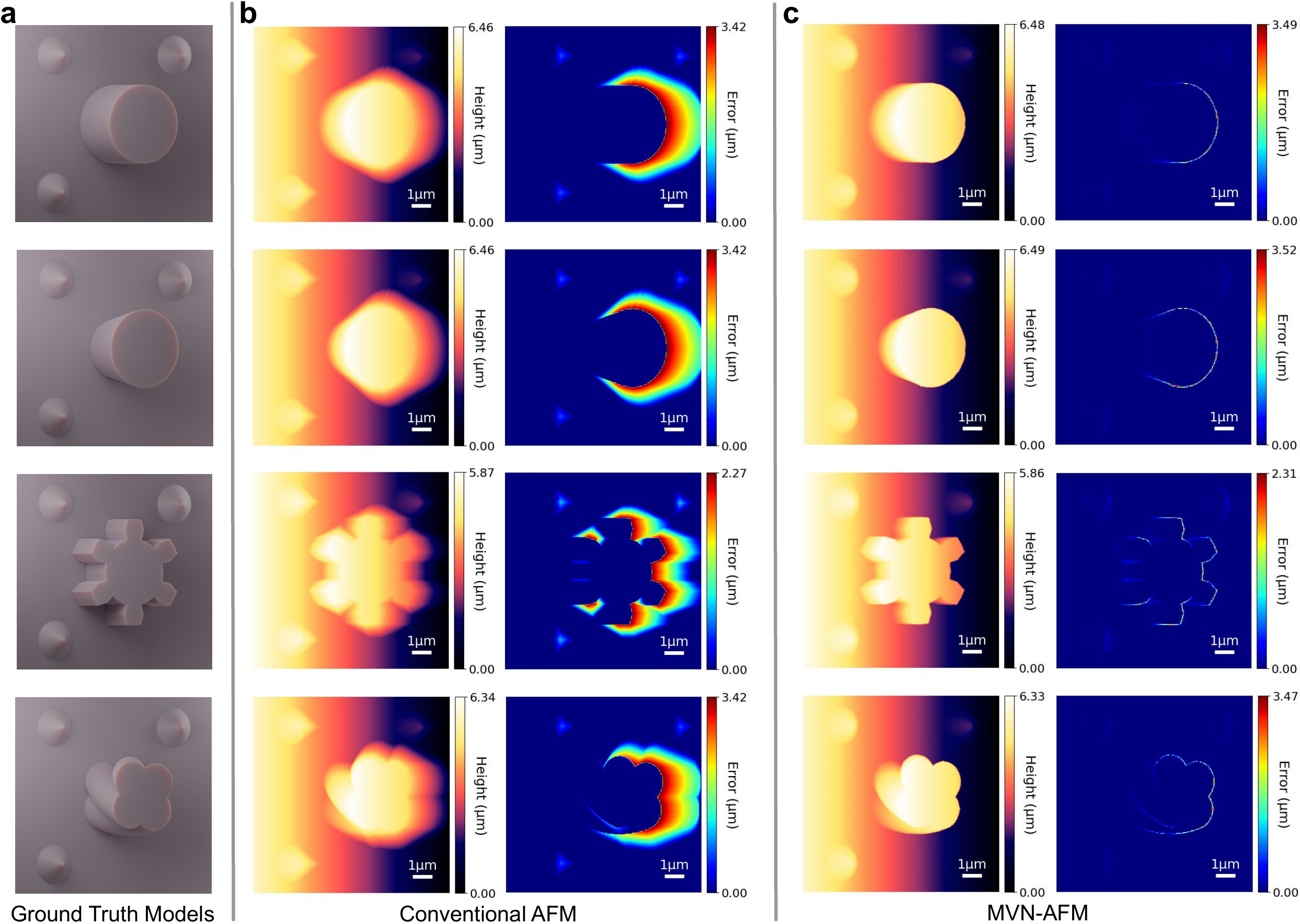}
    \caption[Evaluation of MVN-AFM’s improvement in reconstructed 3D models.]{ 
    \textbf{ Evaluation of MVN-AFM’s improvement in reconstructed 3D models. }
    \textbf{a} The images of ground truth models in the given viewpoints.
    \textbf{b} The simulated AFM images and corresponding height error maps.
    \textbf{c} The topography images of 3D models reconstructed by MVN-AFM and corresponding height error maps.
    }
    \label{fig: more_error_map}
\end{figure*}

\clearpage

\begin{figure*}[h]
    \centering
    \vspace{2em}
    \includegraphics[width=0.9\linewidth]{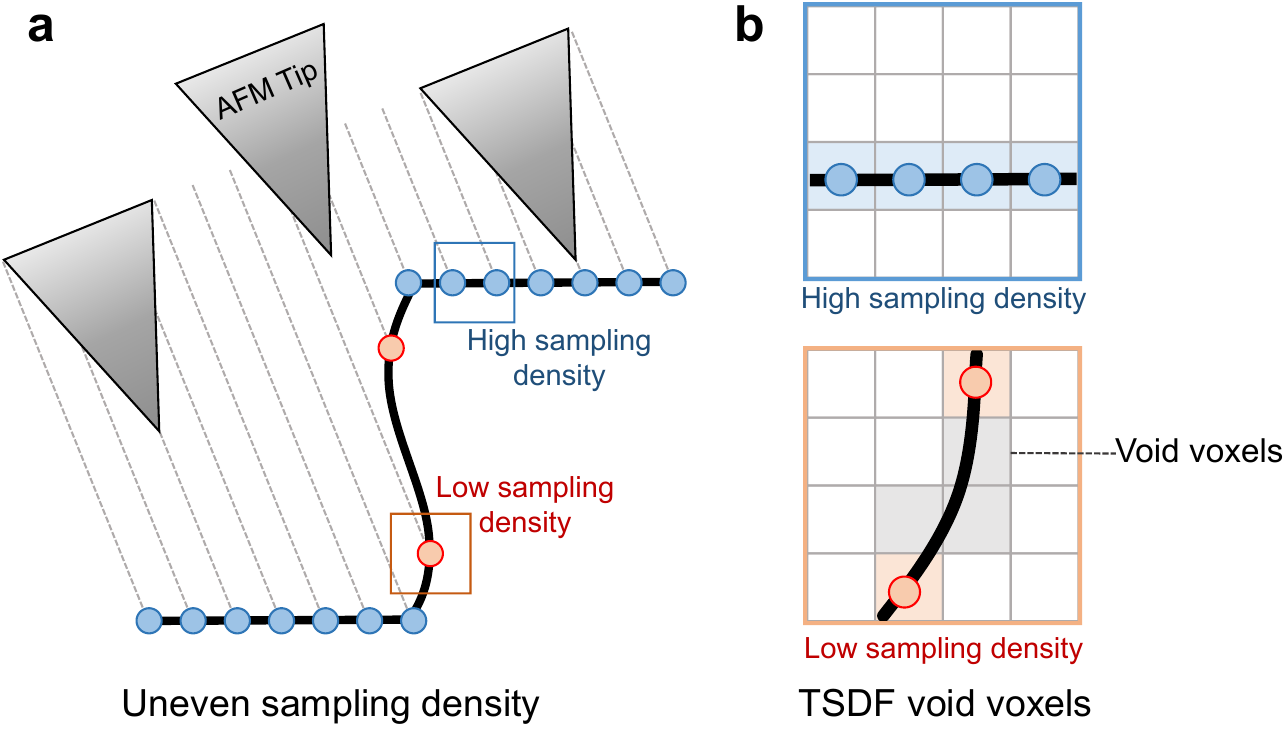}
    \caption[The cause and effect of uneven sampling density in AFM scanning.]{ 
    \textbf{ The cause and effect of uneven sampling density in AFM scanning. }
    \textbf{a} The AFM probe moves in a uniform step on the scanning plane, but the distribution of sample points on the structure surface is uneven due to the structure's shape variations. Because of the restricted range of tilt angles and the limited number of viewpoints in multi-view AFM scanning, the sampling density is relatively low on some surface features. 
    \textbf{b} The TSDF Fusion method divides the space into voxels. However, in regions with very low sampling density, the weights of these voxels have never been updated, resulting in voids in the reconstructed models.}
    \label{fig: uneven_density}
\end{figure*}

\clearpage

\begin{figure*}[h]
    \centering
    \includegraphics[width=0.79\linewidth]{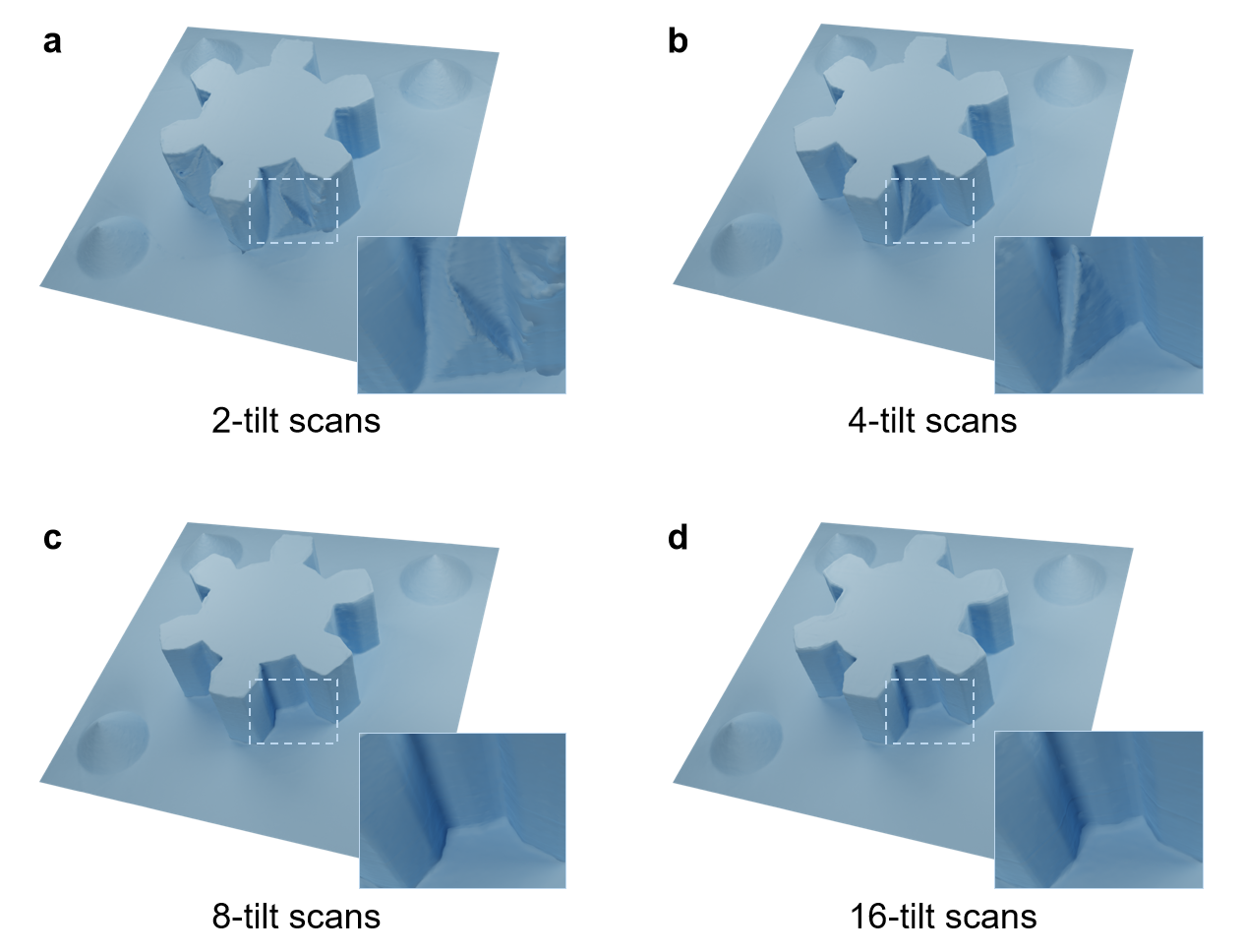}
    \caption[Influence of the number of tilt scans on reconstructed results.]{ 
    \textbf{Influence of the number of tilt scans on reconstructed results.}
    \textbf{a, b, c, d} The gear model reconstructed using 2, 4, 8, and 16 AFM tilt scanning data, respectively.
    \textbf{a} Using only two tilt scanning data as in previous tilting methods for microstructures with complex geometrical features is insufficient.  
    \textbf{b, c, d} With the increased number of tilt views, the AFM scan data provide more geometry information, leading to a more accurate 3D reconstruction of the surface. 
    \textbf{c, d} The visualization of reconstructed models with 8-tilt or 16-tilt scans remains essentially unchanged.
}
    \label{fig: tilt_number}
\end{figure*}

\begin{table*}
  \centering
  \small
  \begin{tabular}{c c c c c c c c | c }
    \toprule
    L1 Chamfer ({\textmu}m) &Cylinder&Undercut&Gear&Spiral&Monkey&House&Average&Time (hour)\\
    \midrule
    2-tilt & 0.0177 & 0.0223 & 0.0364 & 0.0199 & 0.0371 & 0.0537 & 0.0312 & $\approx$0.5\\
    4-tilt & 0.0116 & 0.0120 & 0.0175 & 0.0151 & 0.0238 & 0.0231 & 0.0172 & $\approx$1\\
    8-tilt & 0.0106 & 0.0111 & 0.0133 & 0.0109 & 0.0193 & 0.0189 & 0.0140 & $\approx$2\\
    16-tilt & 0.0107 & 0.0106 & 0.0132 & 0.0118 & 0.0180 & 0.0181 & 0.0137 & $\approx$4\\
    \bottomrule
  \end{tabular}
  \caption[Reconstruction error and the scanning time for different numbers of AFM tilt scans.]{
  \textbf{Reconstruction error and the scanning time for different numbers of AFM tilt scans.}
  The table shows the L1 chamfer distance (lower is better) between the ground truth models and the models reconstructed by different numbers of tilt scans using MVN-AFM.
  The 8-tilt view is the parameter in our experiments.
  The time column demonstrates the approximate time required to complete different numbers of AFM tilt scans in real experiments, including the time for AFM scanning and switching between different viewpoints.
  However, considering the slow speed of AFM scanning, the number of views cannot be increased indefinitely.
  For structures in TPL experiments, there is no considerable reduction in the reconstruction error between 8-tilt and 16-tilt scans, but this would increase the data acquisition time by 2 hours. Therefore, our real-world experiments practiced eight tilt scans for each sample.
  }
  \label{tab:tilt_view_error}
\end{table*}

\end{document}